\documentclass[10pt,twocolumn,letterpaper]{article}

\usepackage{iccv}
\usepackage{times}
\usepackage{epsfig}
\usepackage{graphicx}
\usepackage{amsmath}
\usepackage{amssymb}

\usepackage{multibib}
\newcites{latex}{References}

\usepackage{booktabs}
\usepackage{tabularx}
\RequirePackage{cite}     
\usepackage[pagebackref=true,breaklinks=true,letterpaper=true,colorlinks,bookmarks=false]{hyperref}

\iccvfinalcopy 

\usepackage[capitalize]{cleveref}
\crefname{section}{Sec.}{Secs.}
\Crefname{section}{Section}{Sections}
\Crefname{table}{Table}{Tables}
\crefname{table}{Tab.}{Tabs.}

\def\iccvPaperID{6214} 

\ificcvfinal\fi

\newcommand{\myparagraph}[1]{\smallskip\noindent\textbf{#1}\hspace{0.5em}}

\newcommand{\myparagraphnospace}[1]{\noindent\textbf{#1}\hspace{0.5em}}

\newcommand\blfootnote[1]{%
  \begingroup
  \renewcommand\thefootnote{}\footnote{#1}%
  \addtocounter{footnote}{-1}%
  \endgroup
}

\usepackage{fancyhdr}
\usepackage{setspace}

\fancypagestyle{firststyle}
{

    \fancyhf{}
    \lfoot{{\footnotesize\begin{spacing}{.5}\parbox{\linewidth}{\vspace{3.0em}
    To appear in Proceedings of the \emph{IEEE/CVF International Conference on Computer Vision (ICCV)}, Paris, France, Oct. 2023.%
    \\\hrule\vspace{\baselineskip}
    \copyright~2023 IEEE. Personal use of this material is permitted. Permission from IEEE must be obtained for all other uses, in any current or future media, including reprinting/republishing this material for advertising or promotional purposes, creating new collective works, for resale or redistribution to servers or lists, or reuse of any copyrighted component of this work in other works. 
    }\end{spacing}}}
}

\begin{document}

\title{FunnyBirds: A Synthetic Vision Dataset for a\\ Part-Based Analysis of Explainable AI Methods}

\author{%
  Robin Hesse\textsuperscript{1}
   \and
   Simone Schaub-Meyer\textsuperscript{1,2}
   \and
   Stefan Roth\textsuperscript{1,2} \\
   \and
   \textsuperscript{1}Department of Computer Science, TU Darmstadt \quad \textsuperscript{2}hessian.AI\\
}

\maketitle
\ificcvfinal\fi

\begin{abstract}
The field of explainable artificial intelligence (XAI) aims to uncover the inner workings of complex deep neural models.
While being crucial for safety-critical domains, 
XAI inherently lacks ground-truth explanations, making its automatic evaluation an unsolved problem. 
We address this challenge by proposing a novel synthetic vision dataset, named \emph{FunnyBirds}, and accompanying automatic evaluation protocols.
Our dataset allows performing semantically meaningful image interventions, \eg, removing individual object parts, which has three important implications. First, it enables analyzing explanations on a part level, which is closer to human comprehension than existing methods that evaluate on a pixel level.
Second, by comparing the model output for inputs with removed parts, we can estimate ground-truth part importances that should be reflected in the explanations.
Third, by mapping individual explanations into a common space of part importances, we can analyze a variety of different explanation types in a single common framework.
Using our tools, we report results for 24 different combinations of neural models and XAI methods, demonstrating the strengths and weaknesses of the assessed methods in a fully automatic and systematic manner. \blfootnote{
Dataset and code available at
\href{https://github.com/visinf/funnybirds/}{github.com/visinf/funnybirds/}.} 
\end{abstract}
\thispagestyle{firststyle}

\section{Introduction}
\begin{figure}[t]
  \centering
   \includegraphics[width=0.97\linewidth]{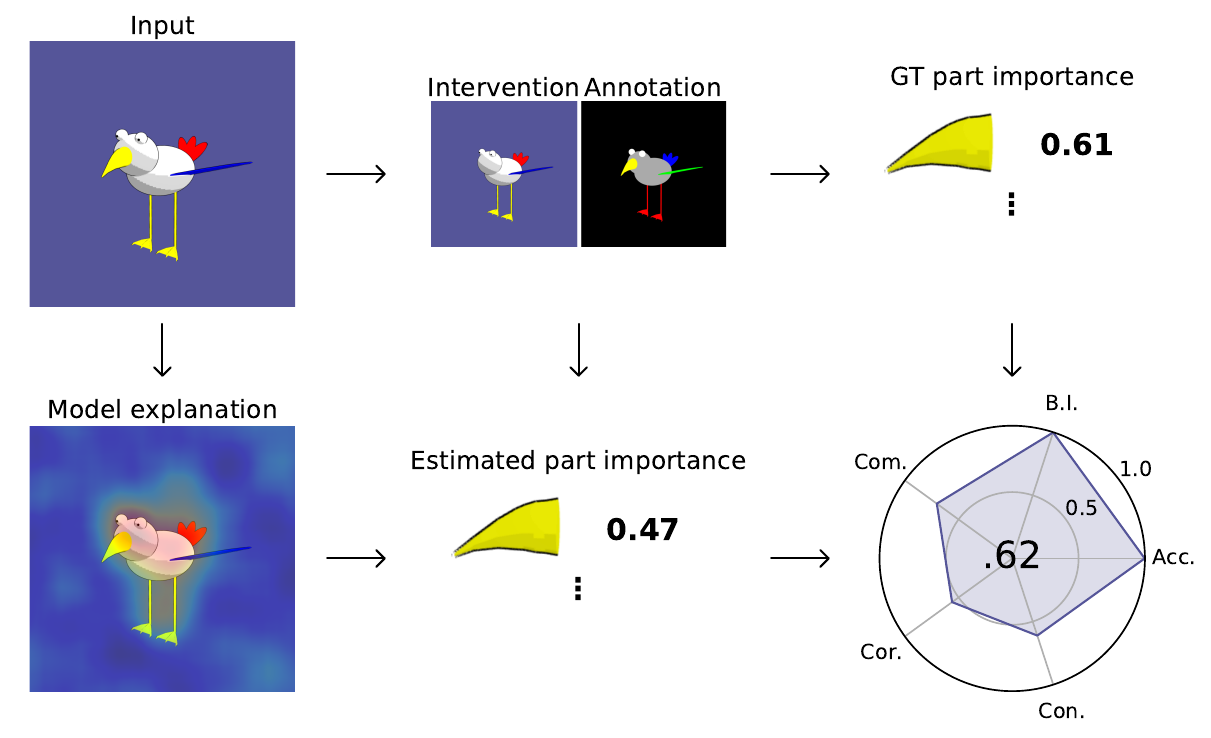}
   \vspace{-1.5mm}
   \caption{\emph{Schematic illustration of our proposed dataset and analysis framework}. For each input image, we render the corresponding part annotation and part interventions (\eg, removed beak). The interventions are used to estimate ground-truth importance scores of each input part, which are then compared to the estimated part importances from the explanation method under inspection (here \cite{Petsiuk:2018:RIS}). We evaluate multiple dimensions of explainability to draw a more conclusive picture. 
   Please refer to \cref{sec:evaluation:benchmark} for details.}
   \label{fig:teaser}
     \vspace{-0.5em}
\end{figure}

Even though deep learning models have achieved breakthrough results in computer vision, their inner workings remain largely opaque.
As a result, deep networks sometimes receive only limited user trust and cannot be applied blindly in safety-critical domains. 
To overcome this issue, a growing interest in the field of explainable artificial intelligence (XAI) has emerged, attempting to explain the inner workings of deep neural models in a human-comprehensible way. However, since there are generally no ground-truth explanations, evaluating XAI methods remains an open challenge.
In fact, a third of XAI papers lack sound quantitative evaluation~\cite{Nauta:2022:FAE}, while other work has limited comparability~\cite{Kim:2022:EHI} or problematic evaluation protocols~\cite{Hase:2021:ODP, Nauta:2022:FAE}.

To overcome the issue of missing ground-truth explanations, automatic evaluations are often done via proxy tasks that adhere to the idea of performing image interventions by removing certain input features to then measure the resulting impact on the model output~\cite{Fong:2017:IEB, Shrikumar:2017:LIF, Hase:2021:ODP}. As image interventions are non-trivial to perform on existing vision datasets, they are usually applied on a pixel level, \eg, masking out single pixels~\cite{Fong:2017:IEB, Shrikumar:2017:LIF, Hesse:2021:FAA, Samek:2017:EVW, Hase:2021:ODP}.
However, this and related approaches share several downsides. 
First, performing interventions, as well as evaluating explanations on a pixel level, is disconnected from the downstream task of providing \emph{human}-understandable explanations since humans perceive images in concepts rather than pixels. Second, existing automatic evaluation protocols are developed for specific explanation types, \eg, pixel-level attribution maps, and thus, cannot be extended to other explanation types like prototypes~\cite{Chen:2019:TLL}. Third, by performing unrealistic interventions in image space, \eg, masking out pixels, they introduce domain shifts compared to the training distribution~\cite{Hase:2021:ODP, Hooker:2019:BIM, Chang:2019:EIC}, which can cause the model to behave unexpectedly, and thus negatively affects the evaluation.

In this work, we address the above and more challenges to contribute an important step toward a more rigorous \emph{quantitative} evaluation of XAI methods by proposing a thorough, dedicated evaluation/analysis tool. We do so by building a \emph{fully controllable}, synthetic classification dataset consisting of renderings of artificial bird species. 
This approach to analyzing XAI methods is analogous to controlled laboratory research, where we have full control over all variables, eliminating the potential influence of irrelevant factors, and therefore, providing clearer evidence of the observed behavior~\cite{Aziz:2017:CBF}.  
Our proposed dataset allows us to make the following main contributions:
\textit{(1)} We cover a wide range of dimensions of explainability by considering a \emph{collection of different evaluation protocols}.
\textit{(2)} We allow to automatically compare various explanation types in a \textit{shared} framework. 
\textit{(3)} We avoid the out-of-domain issue of previous image-space interventions by introducing \emph{semantically meaningful interventions} at training time.
\textit{(4)} We reduce the gap between the downstream task of human comprehension and XAI evaluation by proposing \emph{metrics that operate at a semantically meaningful part level} rather than the semantically less meaningful pixel level. 
 \textit{(5)} We \textit{automatically} analyze the \textit{coherence} of explanations.
\textit{(6)} We analyze 24 different combinations of existing XAI methods and neural models, highlighting their strengths and weaknesses as well as identifying new insights that may be of general interest to the XAI community.

\section{Related Work}
\paragraph{XAI methods.}
In this work, we focus on image classification explanations for the outcome of one particular input using deep neural networks (DNNs).
\emph{Perturbation-based explanation methods}~\cite{Li:2016:UNN, Zeiler:2014:VUC, Zhou:2015:PEN, Zintgraf:2017:VDN, Ribeiro:2016:WSI, Petsiuk:2018:RIS, Abnar:2020:QAF, Chefer:2021:TIB} estimate the impact of pixels/regions by measuring the output change occurring when repeatably perturbing inputs or neurons. \emph{Backpropagation-based explanation methods}~\cite{Selvaraju_2017:GCV, Ancona:2018:TBU, Shrikumar:2017:LIF, Simonyan:2014:DIC, Sundararajan:2017:AAD, Erion:2021:IPD, Hesse:2021:FAA} use importance signals, \eg, gradient information, to estimate the importance of each input feature. Both types of methods generate post-hoc explanations for already trained networks. 
On the other hand, \emph{intrinsically explainable models}~\cite{Boehle:2022:BCN, Bohle:2021:CDA, Nauta:2021:NPT, Melis:2018:TRI, Chen:2019:TLL, Brendel:2019:ACB} consider explainability already in the design process of the model itself. 
For example, B-cos networks~\cite{Boehle:2022:BCN} enforce weight-input alignment to highlight the most important input regions. ProtoPNet~\cite{Chen:2019:TLL} computes the feature similarity between test image patches and prototypes to help explain why a decision was made based on \textit{this} patch looking like \textit{that} prototype.
BagNets\ \cite{Brendel:2019:ACB} use a linear classifier on top of each feature extracted from small image patches to highlight task-important regions.  
$\mathcal{X}$-DNNs\ \cite{Hesse:2021:FAA} remove the bias term in ReLU networks to allow for an efficiently computable closed-form solution of Integrated Gradients~\cite{Sundararajan:2017:AAD}.

\myparagraph{Evaluating XAI methods.} 
A sound evaluation of XAI methods is essential to compare different approaches and unveil their strengths and weaknesses.
Nauta \etal~\cite{Nauta:2022:FAE} argue that explainability is a multi-faceted concept and propose various properties that describe different aspects of explanation quality. Below, we outline existing evaluation protocols for three of these dimensions that have received particularly much attention in related work and that allow for a well-defined automatic evaluation:

\textbf{Correctness} denotes the faithfulness of an explanation \wrt the model~\cite{Nauta:2022:FAE}. 
A common way to measure the correctness of attribution methods is the \textit{Incremental Deletion} protocol~\cite{Fong:2017:IEB, Shrikumar:2017:LIF, Hesse:2021:FAA, Chang:2019:EIC, Hase:2021:ODP, Nauta:2022:FAE, Hooker:2019:BIM, Bohle:2021:CDA, Samek:2017:EVW, Wagner:2019:IFG}, where input pixels are incrementally removed in the order of their relative importance to measure the impact on the output. 
Similarly, in the \textit{Single Deletion} protocol~\cite{Nauta:2022:FAE, Melis:2018:TRI, Chen:2019:ENN, OShaughnessy:2020:GCE, Zhang:2019:ICN}, individual (input) features are deleted to compute the correlation between feature importance and the actual output change. 
A drawback of the above protocols is the domain shift occurring when introducing image interventions, which can cause undesirable model behavior influencing the metrics~\cite{Hase:2021:ODP, Hooker:2019:BIM, Chang:2019:EIC}. To reduce this, masked pixels can be inpainted with a generative model~\cite{Chang:2019:EIC} or text classification datasets can be augmented with training inputs that have similar alterations as the evaluation protocol~\cite{Hase:2021:ODP}. However, none of these approaches achieve an \emph{exact} alignment of the train and test domain as we do in this work.

\textbf{Completeness} measures the degree to which an explanation describes all aspects of the model's decision~\cite{Nauta:2022:FAE}. 
In the \textit{Preservation/Deletion Check} protocols~\cite{Nauta:2022:FAE, Chang:2019:EIC, Chen:2018:LEI, Dhurandhar:2018:EBM, Pope:2019:EMG}, important features are preserved/deleted to test if the output prediction stays the same/changes. However, as above, these protocols introduce unrealistic domain changes. In the \textit{Controlled Synthetic Data Check} protocol~\cite{Nauta:2022:FAE, Oramas:2019:VEI, Adebayo:2020:DTM, Kim:2018:IBF, Ross:2017:RRR, Chen:2018:LEI}, a synthetic dataset is used to evaluate if the explanation of a model with almost perfect accuracy aligns with the dataset generation process. 
While closely related to our work, existing synthetic datasets for evaluating XAI methods are very small-scale, only evaluate a single dimension of explainability, and do not allow for diverse image-space interventions that are crucial for evaluating \emph{correctness}.

\textbf{Contrastivity} measures the degree to which an explanation is discriminative \wrt different outputs~\cite{Nauta:2022:FAE}. It can be measured with the \textit{Target Sensitivity} protocol~\cite{Nauta:2022:FAE} that evaluates if the explanations for two different targets are different~\cite{Pope:2019:EMG} and sometimes, even more, whether the explanation aligns with the image region of the correct target~\cite{ Zhang:2016:TDN, Bohle:2021:CDA, Du:2018:TED}. 
However, these evaluations can require accurate object annotations, which are not given for most classification datasets, or they can also introduce out-of-domain issues when arranging images in a grid as in \cite{Bohle:2021:CDA, Rao:2022:TBU, Boehle:2022:BCN}.

\myparagraph{Human studies.} Besides above \emph{automatic} evaluations, there have been efforts to assess the quality of XAI methods with \emph{human} subjects~\cite{Shen:2020:HUM, Nguyen:2021:EFA, Russakovsky:2015:ILS, Colin:2021:WIC}. 
Human studies nicely reflect the actual downstream tasks of explanations, \ie, helping \emph{humans} to understand models, and they can be applied to explanations that go beyond simple heat maps~\cite{Kim:2022:EHI}. Downsides of user studies include larger cost and human biases~\cite{McCambridge:2012:EDC}. 
Further, humans are not capable of directly evaluating if explanations are correct \wrt a complex, deep model. 
With our custom analysis, we aim to reduce the gap between the automatic evaluation and human studies while addressing the described drawbacks of human assessment.

\myparagraph{Synthetic datasets.} 
Synthetic datasets have an important and long history in many other disciplines of computer vision~\cite{Richter:2016:PDG, Dosovitskiy:2015:FNL, Butler:2012:NOS, Johnson:2017:DDC, Park:2019:RCC}. Richter \etal~\cite{Richter:2016:PDG} utilize a video game to create a large-scale segmentation dataset.
The ``Flying Chairs'' dataset~\cite{Dosovitskiy:2015:FNL} and MPI Sintel~\cite{Butler:2012:NOS} are popular synthetic datasets used for the training and evaluation of optical flow methods. CLEVR~\cite{Johnson:2017:DDC} was introduced to test different visual reasoning abilities. All these works advanced their specific sub-fields, showing that synthetic datasets are a valuable tool, even in the absence of full realism.

\section{The FunnyBirds Dataset}\label{sec:dataset}
We propose a synthetic vision dataset that is developed to automatically and quantitatively analyze XAI methods. 
To this end, we made a multitude of carefully considered
design choices, which we summarize in the following.

\begin{figure}[t]
  \centering
   \includegraphics[width=1\linewidth]{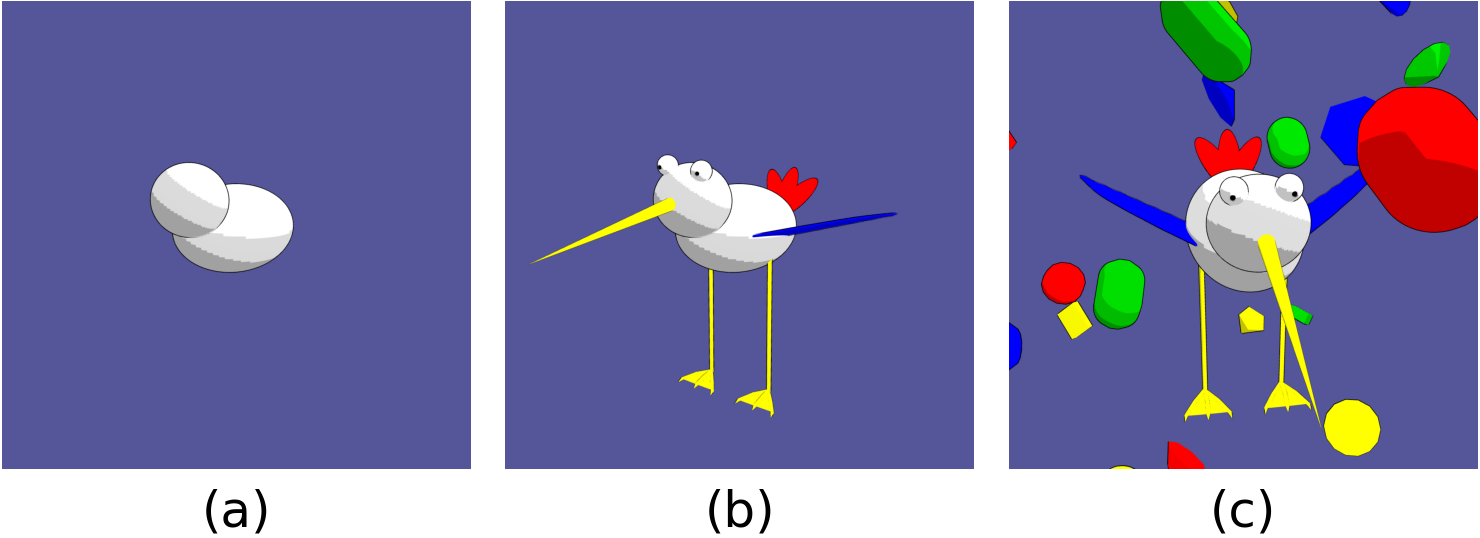}
   \vspace{-12pt}
   \caption{\textit{Generation process of our synthetic dataset.} \textit{(a)} The scene is initialized with a neutral bird body part. \textit{(b)} Class-specific parts are added. \textit{(c)} To reflect real-world challenges, random backgrounds, illumination, and viewpoint changes are added/applied.}
   \label{fig:generation}
   \vspace{-0.5em}
\end{figure}

Since XAI methods in computer vision have been developed mainly in the context of classification to date~\cite{Nauta:2022:FAE}, we focus on classification and propose a fine-grained bird species dataset inspired by the CUB-200-2011 dataset~\cite{Wah:2011:CUB} commonly used by XAI methods~\cite{Nauta:2021:NPT, Chen:2019:TLL, Zhang:2018:ICN, Hendricks:2016:GVE, Goyal:2019:CVE}.
Our dataset consists of 50\,500 images (50k train, 500 test) of 50 synthetic bird species.
We found 500 test images to be sufficient to produce stable results while allowing efficient evaluations with limited hardware (see supplement).

One particularly important desideratum behind our design
process is the notion of \textit{concepts} that we define as mental representations of psychological entities that are crucial to categorization~\cite{Margolis:2022:C, Rips:2012:CCM, Zentall:2002:CCL, Meding:1999:CTC}. 
For instance, humans might categorize a bird as a flamingo when observing the concepts curved \textit{beak} and pink \textit{wings}. Similarly, they would communicate their reasoning process in terms of concepts~\cite{Rips:2012:CCM}.
Hence, we argue that concepts are essential in the context of \textit{XAI for humans}, and thus, we design our dataset with a particular emphasis on the notion of concepts. 
To derive a precise level of granularity for the concepts in our dataset, we enforce them to be as granular as possible, while corresponding to an existing individual word, and being attached to the bird's body. This avoids ``unrealistic'' scenarios where the removal of one concept leaves another concept unattached (the body of the bird is never removed) and eliminates overly fine-grained concepts. Following these rules, our FunnyBirds consist of the five human comprehensible concepts \textit{beak}, \textit{wings}, \textit{feet}, \textit{eyes}, and \textit{tail}, to which we will refer as \textit{parts} in the following.

We manually design $4$ beaks, $3$ eyes, $4$ feet, $9$ tails, and $6$ wings of predefined varied shape and/or color (see supplement). 
Every FunnyBird class consists of a unique combination of these parts. 
We randomly sample 50 classes out of the 2\,592 combinations as our dataset.
Using these classes, we follow the data-generation process outlined in \cref{fig:generation} by adding class-specific parts to a neutral body model to obtain an instance of a FunnyBird.  
To reflect real-world challenges, we randomly add background objects and change the illumination and viewpoint. 
We also explored more photo-realistic simulations with real background images and textured parts. However, with neural models learning to ignore these ``distractions'', our quantitative analysis was largely unaffected. We thus omit this in pursuit of simplicity. 
We verified the solvability of our dataset with a human expert who achieved an accuracy of $97.4\%$ on the proposed test set.
To ensure that removing any part does not result in another valid class, each complete FunnyBird instance contains all five parts.
To ensure that images with removed parts are not out-of-distribution, we include images with missing parts in the training set.
As these images can no longer be associated with \emph{one} specific class, we use a multi-label classification training scheme where all classes that contain all of the remaining parts are considered valid (with equal probability). 

Our data-generation process has several important implications that make it particularly well suited for evaluating and analyzing XAI methods. First, based on the assumption that humans think in concepts rather than pixels, we argue that a concept-based dataset and the resulting evaluations are better aligned with downstream tasks of human comprehension than pixel-based evaluations. Second, we have pixel-accurate annotations of each object and bird part in the scene, which is required for various automatic evaluation protocols and tailored analyses. Third, having full control over the data-generation process allows applying realistic image-space interventions, such as removing parts, which can be used to estimate ground-truth importance scores.

\section{Evaluation Protocols}
\label{sec:evaluation}
We will now outline how our dataset and its functionalities can be used for the quantitative evaluation and analysis of XAI methods. 
We first propose a \emph{general, multi-dimensional analysis framework}, named FunnyBirds framework, that allows to evaluate a wide range of existing XAI methods in a common framework. Second, we propose a more tailored way of analysis where our dataset's capabilities are used to gain deeper \emph{insights into specific methods} similar to what human studies could do.

\subsection{The FunnyBirds framework}\label{sec:evaluation:benchmark}

The multi-dimensional FunnyBirds analysis framework for XAI methods entails six evaluation protocols for three explainability dimensions, \ie, \textit{completeness}, \textit{correctness}, and \textit{contrastivity}. They were chosen for their popularity in related work~\cite{Nauta:2022:FAE}, ability to be automatically measurable, and compatibility with our dataset. 
Most of our protocols are inspired by well-established evaluation practices and follow accepted assumptions. 
Our dataset generation process ensures that all conducted interventions can be considered in-domain and semantically meaningful, eliminating common downsides~\cite{Hase:2021:ODP, Hooker:2019:BIM} of several existing evaluation protocols that rely on image interventions~\cite{Fong:2017:IEB, Shrikumar:2017:LIF, Hesse:2021:FAA, Chang:2019:EIC, Hase:2021:ODP}. 

\myparagraph{Meaning of \textit{ground truth}.} Our work relies on interventions to estimate ``ground-truth'' importance scores for individual parts. This is motivated by \emph{(1)} the observation that in XAI we aim to approximate the causal structure of the inspected model, and \emph{(2)} findings that the causal structure can be estimated via interventions~\cite{Peters:2017:ECI}. Since the perfect causal structure (which is too complex to be comprehended) is already given by the model itself, we emphasize that a \textit{simplified approximation} is needed for XAI, instead. We thus assume such an approximation for estimating the ground-truth scores with interventions. For example, in the single deletion protocol (see \cref{sec:correctness}), we assume that a hypothetical \textit{simplified} model processes each part individually, which generally does not need to hold for the \textit{real} model.
This assumption is implicitly also made in existing single deletion protocols~\cite{Melis:2018:TRI, Chen:2019:ENN, OShaughnessy:2020:GCE, Zhang:2019:ICN}. Further, most existing explanation methods, \eg, attribution maps, communicate the importance of each feature \textit{individually}, thus evaluating explanations beyond individual concepts is not yet sensible. 

\myparagraph{Interface functions.}
The diversity of explanation types presents a major challenge to evaluating XAI methods, which prevents a direct comparison of different methods and created a need for various non-standardized evaluations. We thus propose to build our evaluation framework around so-called \textit{interface functions} that implement more general properties, such as which object parts are important. They can be derived from a variety of explanation types, and thus, used to evaluate distinct explanations in a common framework. This allows us to not only compare a spectrum of existing methods, but also to potentially include future, novel explanation types. Our evaluation framework currently uses the following interface functions:
\begin{description}
  \item[$\textnormal{PI}(\cdot)$] -- The part importance score for each part. For example, the summed attribution within that part.
  \item[$\textnormal{P}(\cdot)$] -- The parts that are considered to be important, derived from the output of an XAI method. For attribution methods, a part can be considered important if its part importance is at least $t\%$ of the total attribution.
\end{description}
Note that different explanation types vary in their suitability for differing downstream tasks, and thus, also in their suitability for above interface functions. 
Furthermore, just as humans differ in interpreting various explanations, there may not be a single correct way of implementing the interface functions (see supplement). Therefore, comparing different explanation types is not absolutely fair and should be done with caution. Nonetheless, we argue that estimating important image regions and their relative importance are fairly basic challenges in XAI, and thus, appropriate downstream tasks, respectively, interface functions.

\myparagraph{Framework.}
For a simple comparison and visualization, we design all metrics to be in the range $[0,1]$, with higher values indicating better scores. 
For each dimension (as described below), we report a single score comprised of all included protocols (see \cref{tab:protocols}). 
Additionally, we report the average of all our explanation quality dimensions, \ie, \textit{completeness}, \textit{correctness}, \textit{contrastivity}, as a single score that summarizes the overall quality of an XAI method for the inspected model. We let $\mathcal{D}$ denote our dataset of size $N$, consisting of images $x_n$ with class label $c_n$; $f$ is the inspected model with $f(x_n)$ denoting the predicted class or the logit of the target class, and $e_f(x_n)$ is the explanation of $f(x_n)$ \wrt the target class $c_n$.

\begin{table*}
  \centering
  \caption{\textit{Overview of our evaluation protocols.} ``Interface function'' denotes the required interface function for each protocol.}
  \smallskip
  \small
  \begin{tabularx}{\textwidth}{@{}lllX@{}}
    \toprule
    Dimension & Protocol & Interface function & Description\\
    \midrule
    -- & A & -- & Percentage of correct predictions over the test dataset \\
    -- & BI & -- & The model should not be sensitive to the background objects  \\
    Completeness & CSDC & $\text{P}$ & Explanation should highlight all relevant parts  \\
    Completeness & PC & $\text{P}$ & Preserving parts identified as important should result in the same prediction\\
    Completeness & DC & $\text{P}$ & Removing parts identified as important should result in a different prediction\\
    (Over-)Completeness & D & $\text{P}$ & Explanation should not highlight irrelevant parts \\
    Correctness & SD & $\text{PI}$ & Estimated importance of each part should be correlated to actual importance\\
    Contrastivity & TS & $\text{PI}$ & Explanations for different classes should highlight class-specific parts\\
    \bottomrule
  \end{tabularx}
  \label{tab:protocols}
  \vspace{-0.5em}
\end{table*}

\subsubsection{Accuracy and background independence}
\paragraph{Accuracy (A).} An overly simple model may be explainable but not solve the task at hand. To detect such cases, our framework reports the standard classification accuracy.

\myparagraph{Background independence (BI).} Similarly, another trivial solution to achieve high explanation scores is a model that is sensitive to the entire image. An explanation would consequently highlight the entire image. To detect this case, we report the background independence as the ratio of background objects that are unimportant, \ie, that, when removed, cause the target logit to drop less than $5\%$.

\subsubsection{Completeness}\label{sec:evaluation:com}

\paragraph{Controlled synthetic data check (CSDC).}
Our transparent data generation process allows us to know the subsets of parts $\{\mathcal{P}^\prime_{c_n,i} \mid i=1,\ldots\}$ that are sufficient to correctly categorize an image of class $c_n$ containing parts $\mathcal{P}_{c_n}$. A model that accurately captures the essence of our dataset, \ie, has an accuracy close to $1.0$, must thus consider at least one of these sets $\mathcal{P}^\prime_{c_n,i}$~\cite{Nauta:2022:FAE, Oramas:2019:VEI, Ross:2017:RRR, Chen:2018:LEI}. To ensure that this assumption holds and CSDC is faithful \wrt the model, we exclude misclassified samples from this protocol. 
We measure the CSDC completeness of an explanation as the set overlap between the parts that are estimated to be important by the explanation (\ie, $\text{P}(e_f(x_n))$ where $\text{P}(\cdot)$ is the interface function giving the important parts) and the sufficient part set that has the largest overlap with $\text{P}(e_f(x_n))$, normalized by the size of the sufficient part set:
\begin{equation}
\begin{aligned}
\text{CSDC} =  \frac{1}{N} \sum_{n=1}^{N} \underset{i}{\max} \  \frac{\big|\text{P}(e_f(x_n)) \cap \mathcal{P^\prime}_{c_n,i}\big|}{|\mathcal{P^\prime}_{c_n,i}|}\ .
\end{aligned}
\end{equation}

\myparagraph{Preservation check (PC).}
If an explanation is complete, preserving only the parts of the input that are estimated to be important by the explanation should still result in the same classification output~\cite{Nauta:2022:FAE, Chen:2018:LEI, Dhurandhar:2018:EBM}. To verify this, we first pass the original image through the inspected model to obtain the classification and the corresponding explanation. Afterward, we delete all parts of the image that are estimated to \emph{not} be important by the explanation, \ie, the inverse of $\text{P}(\cdot)$, and measure how often the class prediction remains unchanged. We define our PC score as
\begin{equation}
\text{PC} = \frac{1}{N} \sum_{n=1}^{N} \big[f(x_n^\prime) = f(x_n) \big]\ ,
\label{eq:protocol_pc}
\end{equation}
with $[\cdot]$ denoting the Iverson bracket~\cite{Knuth:1992:TNN} and $x_n^\prime$ denoting the image arising by removing all bird parts but $\text{P}(e_f(x_n))$ from $x_n$. \Ie, the parts estimated to be important remain.

\myparagraph{Deletion check (DC).}
Similarly, if an explanation is complete, deleting all the parts of the input that are estimated to be important by the explanation should result in a different classification output~\cite{Nauta:2022:FAE, Pope:2019:EMG, Dhurandhar:2018:EBM}. Thus, the deletion check is the inverse of PC and defined as
\begin{equation}
\text{DC} = \frac{1}{N} \sum_{n=1}^{N} \big[f(x_n^{\prime\prime}) \neq f(x_n) \big]\ ,
\label{eq:protocol_dc}
\end{equation}
with $x_n^{\prime\prime}$ denoting the image arising by removing the estimated important parts $\text{P}(e_f(x_n))$ from $x_n$.

\myparagraph{Distractibility (D).} 
As over-complete explanations would achieve high completeness scores, we argue that completeness should be considered together with a metric that counters over-completeness (similar to precision and recall).
To this end, we introduce a novel over-completeness score that measures how many actually unimportant input parts are also estimated to be unimportant by the explanation. 
In detail, we successively remove each background object and bird part from the input and consider it to be unimportant if the logit target output drops less than $5\%$ compared to the output for the original image. Next, we measure the overlap between parts $\text{P}(e_f(x_n))$ estimated to be important by the explanation and parts $\mathcal{P}^{\prime \prime}_{f(x_n)}$ that are actually unimportant, normalized by the number of unimportant parts. As the overlap should be small for good evaluations, we subtract the result from 1: 
\begin{equation}
\begin{aligned}
\text{D} = 1 - \frac{1}{N} \sum_{n=1}^{N} \frac{\big|\text{P}(e_f(x_n)) \cap \mathcal{P}^{\prime \prime}_{f(x_n)}\big|}{|\mathcal{P}^{\prime \prime}_{f(x_n)}|}\ .
\end{aligned}
\label{eq:protocol_d}
\end{equation}

The final \textit{completeness score} (Com.) is computed as the mean of the averaged three completeness metrics ($\text{CSDC}$, $\text{PC}$, and $\text{DC}$), and the over-completeness $\text{D}$. 
By having complementing metrics, we can also automatically estimate the threshold $t$ of $\text{P}(\cdot)$, such that $t$ optimally trades off the completeness and over-completeness metrics.

\subsubsection{Correctness}\label{sec:correctness}

\paragraph{Single deletion (SD).}
In the single deletion protocol, we measure the rank correlation between the explanation’s importance score of each part and the actual amount that the model's logit of the target class changes when removing that part~\cite{Nauta:2022:FAE, Melis:2018:TRI, Chen:2019:ENN}. Specifically, we define $\text{SD}$ as
\begin{equation}
\text{SD} = \frac{1}{2} + \frac{1}{2N} \sum_{n=1}^{N} \rho\big(\text{PI}(e_f(x_{n})), f(x_{n}) - \{f(x_{n}^{\prime\prime\prime})\}\big),
\end{equation}
with $\rho$ denoting the Spearman rank-order correlation coefficient, $\text{PI}(\cdot)$ the part importance interface function, and $\{f(x_{n}^{\prime\prime\prime})\}$ the target scores of the images obtained by removing any individual bird part from $x_n$. 

\subsubsection{Contrastivity}

\paragraph{Target sensitivity (TS).} If an explanation is sensitive to a target class, it should highlight image regions that are relevant for the respective target~\cite{Nauta:2022:FAE, Bohle:2021:CDA, Boehle:2022:BCN}. To measure target sensitivity, we take an input sample and select two classes $\hat{c}_1$ and $\hat{c}_2$ that each have exactly two non-overlapping common parts with the actual class of input sample $x_n$. We choose two parts because this is the floor of half the 5 bird parts available in each image. Afterward, we compute the explanations $e_f(x_n, \hat{c}_i)$ with respect to $\hat{c}_1$ and $\hat{c}_2$ and evaluate if they correctly highlight the parts belonging to each respective class: 
\begin{equation}
\begin{aligned}
\text{TS} =
\frac{1}{2N} \sum_{n=1}^{N} & \big[\text{PI}^\prime(e_f(x_n, \hat{c}_1)) > \text{PI}^\prime(e_f(x_n, \hat{c}_2))\big]\\ + &\big[\text{PI}^{\prime\prime}(e_f(x_n, \hat{c}_1)) < \text{PI}^{\prime\prime}(e_f(x_n, \hat{c}_2))\big] \ ,
\end{aligned}
\end{equation}
with $\text{PI}^\prime(\cdot)$ and $\text{PI}^{\prime\prime}(\cdot)$ denoting the summed part importan\-ces of the two parts belonging to class $\hat{c}_1$, respectively $\hat{c}_2$. To ensure that our protocol is faithful \wrt the model, we only consider images where removing parts belonging to $\hat{c}_1$ causes a larger model output drop for $\hat{c}_1$ than removing parts belonging to $\hat{c}_2$, and vice versa.

\begin{figure*}[t!]
  \centering
   \includegraphics[width=1\linewidth]{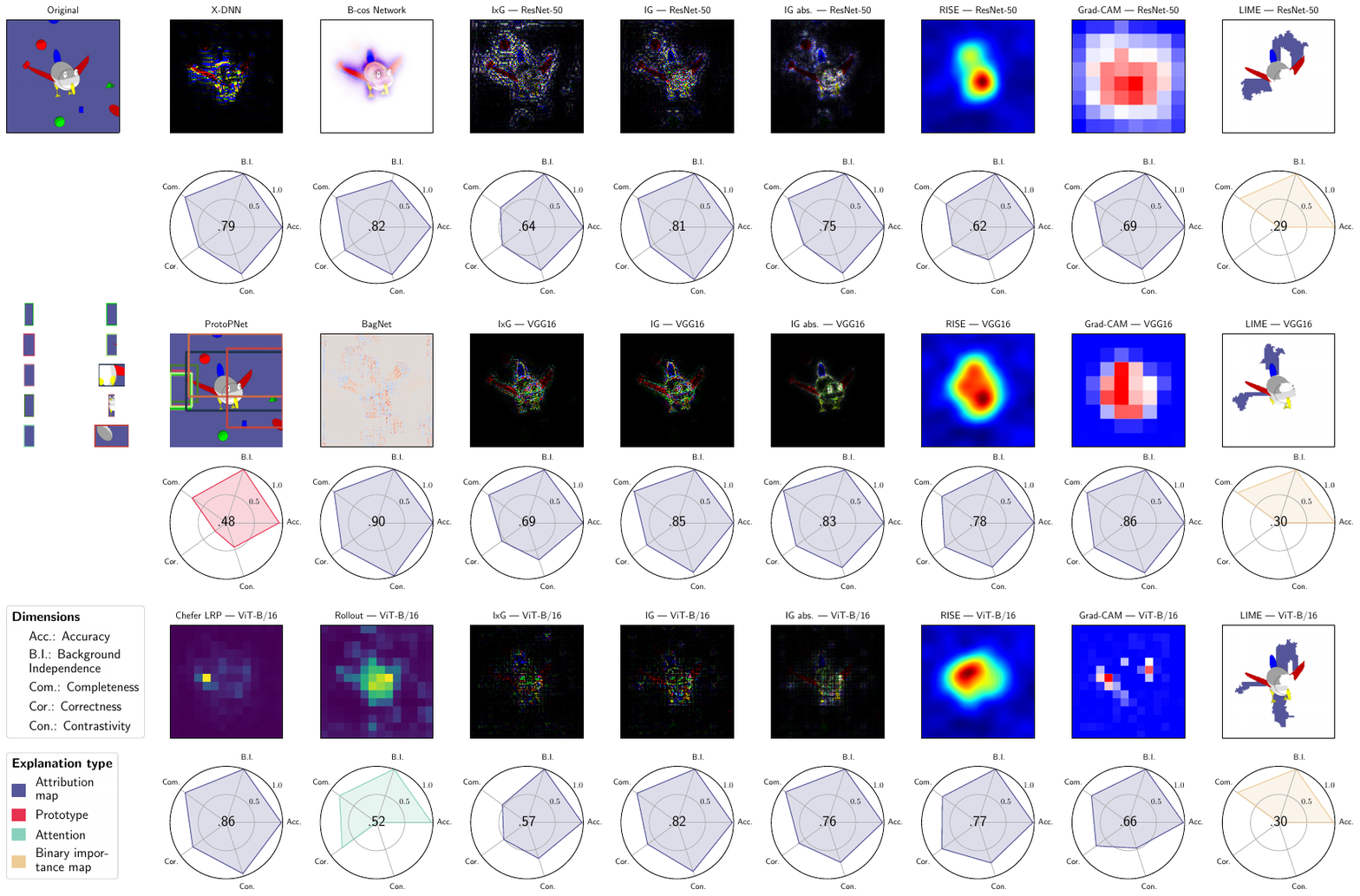}
   \vspace{-20pt}

   \caption{\emph{FunnyBirds evaluation results.} Qualitative and quantitative results for our FunnyBirds framework for the XAI methods Input$\times$Gradient (IxG)~\cite{Shrikumar:2016:NJB}, (absolute) Integrated Gradients (IG (abs.))~\cite{Sundararajan:2017:AAD}, Grad-CAM~\cite{Selvaraju_2017:GCV, Chefer:2021:TIB}, RISE~\cite{Petsiuk:2018:RIS}, LIME~\cite{Ribeiro:2016:WSI}, Rollout~\cite{Abnar:2020:QAF}, Chefer LRP~\cite{Chefer:2021:TIB}, $\mathcal{X}$-DNN~\cite{Hesse:2021:FAA}, BagNet~\cite{Brendel:2019:ACB}, ProtoPNet~\cite{Chen:2019:TLL}, and B-cos network~\cite{Boehle:2022:BCN}. Model-agnostic methods are evaluated on VGG16~\cite{Simonyan:2015:VDC}, ResNet-50~\cite{He:2016:DRL}, and ViT-B/16~\cite{Dosovitskiy:2021:IWW}. Other models use a ResNet-50 backbone, if applicable. Quantitative results are averaged over the entire test set; qualitative results are shown for a representative example. Chart colors indicate different explanation types that should only be compared with caution.
   The center score denotes the mean of the completeness (Com.), correctness (Cor.), and contrastivity (Con.) dimensions. Additionally, we report the accuracy (Acc.) and background independence (B.I.). See \cref{sec:evaluation:benchmark} for details.}
   \label{fig:results}
   \vspace{-0.5em}
\end{figure*}

\subsection{The FunnyBirds custom analysis}
\label{sec:evaluation:custom}
As it is almost impossible to analyze all aspects of an explanation with a single framework, we also show how full control over our dataset enables more tailored and systematic \textit{custom} evaluations.
Similar to recent human studies~\cite{Kim:2022:EHI}, we evaluate the \emph{coherence} of the assessed methods, \ie, how reasonable an explanation is for a human. To the best of our knowledge, this is one of the first methods for evaluating coherence in a fully \textit{automatic and systematic} manner. We give two illustrative examples for implementing such an evaluation, but other instantiations are possible.

\myparagraph{ProtoPNet.} ProtoPNet~\cite{Chen:2019:TLL} promotes explainability by communicating which patches of the training data, called prototypes, look similar to patches in a test image.
However, this only increases interpretability if the detected similarities are compatible with human notions of similarity. To quantitatively verify this in an automatic fashion, we test if relevant part types (\eg, beak or wing) in the detected patch of the test image are also contained in the corresponding prototype identified by ProtoPNet. This assumes that different part types, \ie, beak and wings, do not look similar to most humans, and therefore, are not helpful similarities for explainability. While debatable for real images, we argue that this is a reasonable assumption for our dataset. 

\myparagraph{Counterfactual visual explanations.} 
The objective of counterfactual visual explanations (CVE)~\cite{Goyal:2019:CVE} is to highlight image regions that discriminate the class $c$ of a query image from the class $c^\prime$ of a distractor image. To this end, CVE extracts convolutional neural network (CNN) feature maps from the query  and distractor images to then perform feature swaps between the two feature maps that, when fed to the CNN classifier, lead to the strongest class change (from $c$ to $c^\prime$). 
The corresponding image locations of the feature swaps are then highlighted to yield the CVE.
We argue that this kind of explanation is only meaningful for humans if the swaps are semantically correct, \ie, when a beak is swapped with another beak but not a foot. 
Using our proposed dataset, we conduct a systematic study and
measure the average number of swaps that do not touch the same bird part in both images, and thus, can be considered semantically incorrect. We evaluate four different setups of increasing difficulty where we only use aligned, respectively unaligned, viewpoint pairs with classes that are different in only one, respectively multiple parts.

\myparagraph{Limitations.} 
As with controlled laboratory research~\cite{Aziz:2017:CBF}, our dataset does not reflect all real-world challenges, and thus, is not fully representative of natural images. However, we argue that, as long as XAI methods are challenged to truthfully explain models trained even on our controlled dataset, our proposed method is a valuable tool to faithfully assess the quality of XAI methods.
Nonetheless, for future XAI methods we suggest a two-stream evaluation that utilizes our proposed framework and custom analysis for theoretically grounded assessments of the explanation quality \emph{in addition} to benchmarks that work on natural image data, \eg, Quantus~\cite{Hedstrom:2022:QEA}.
Further, while the dimensions of explainability evaluated in our work can be considered relevant for many applications, there are dimensions that we omit here (see supplement). However, our framework can easily be extended with additional protocols if needed.

\section{Results and Interpretation}
\subsection{FunnyBirds framework}
Results for our multi-dimensional FunnyBirds framework across a broad variety of different XAI methods and three different backbones are given in \cref{fig:results}. For detailed results of each individual evaluation protocol, please refer to the supplement. We plot qualitative explanations for a random image and give the evaluation scores from \cref{sec:evaluation:benchmark} averaged over the entire test set. From this, we can make several interesting findings and derive various hypotheses: 

\myparagraphnospace{(1)} Overall, BagNet~\cite{Brendel:2019:ACB} achieves the highest explainability score, which could be due to its internal model structure that is better aligned with the explanatory power of an attribution map, \ie, pixels are processed more independently.

\myparagraphnospace{(2)} On average, Integrated Gradients (IG)~\cite{Sundararajan:2017:AAD} performs best among model-agnostic methods for different backbones, which is in line with the literature when considering gradient-based approaches~\cite{Shrikumar:2016:NJB, Sundararajan:2017:AAD}. However, LIME~\cite{Ribeiro:2016:WSI} and IG achieve similar completeness, with LIME lacking the ability to communicate meaningful importance scores, which impairs practicability for certain downstream tasks. 
Interestingly, taking the absolute value of IG impairs the mean explanation score slightly, contradicting findings that have been made on pixel-level protocols~\cite{Yang:2023:RFA}. Specifically, it slightly increases completeness while reducing contrastivity. We hypothesize that the negative gradient information is therefore particularly important in identifying regions that do \textit{not} belong to the class of interest. Removing the bias term ($\mathcal{X}$-DNN~\cite{Hesse:2021:FAA} \vs ResNet-50~\cite{He:2016:DRL} w/ IG) slightly improves completeness and correctness while reducing contrastivity.

\myparagraphnospace{(3)} 
Explanation scores for model-agnostic XAI methods are consistently better for VGG16~\cite{Simonyan:2015:VDC} than for ResNet-50~\cite{He:2016:DRL}, indicating that different backbones vary in their explainability. 
We also observe that the relative ordering of different XAI methods changes across different backbones (\eg, RISE~\cite{Petsiuk:2018:RIS} and IxG~\cite{Shrikumar:2016:NJB}), showing the importance of evaluating future XAI methods on different backbones.

\myparagraphnospace{(4)} IxG~\cite{Shrikumar:2016:NJB} and Grad-CAM~\cite{Selvaraju_2017:GCV, Chefer:2021:TIB} seem to work better on CNN-based architectures~\cite{Simonyan:2015:VDC, He:2016:DRL}. 
Methods developed for vision transformers, \ie, Rollout~\cite{Abnar:2020:QAF} and Chefer LRP~\cite{Chefer:2021:TIB}, provide strong explanations, with Chefer LRP achieving the second highest explanation score.
Rollout cannot compute class-specific explanations, which could be disadvantageous in certain downstream tasks.

\myparagraphnospace{(5)} Coarser methods like Grad-CAM~\cite{Selvaraju_2017:GCV} and RISE~\cite{Petsiuk:2018:RIS} tend to be more incomplete, which could be an artifact of poor localization~\cite{Rao:2022:TBU} and our parts being fairly small.

\myparagraphnospace{(6)} The B-cos network~\cite{Boehle:2022:BCN} outperforms competing methods on the same backbone, showing an improved explainability, which is in line with results from~\cite{Boehle:2022:BCN}. 
Surprisingly, we see a slight decrease in background independence.

\begin{figure}[t]
  \centering
   \includegraphics[width=1\linewidth]{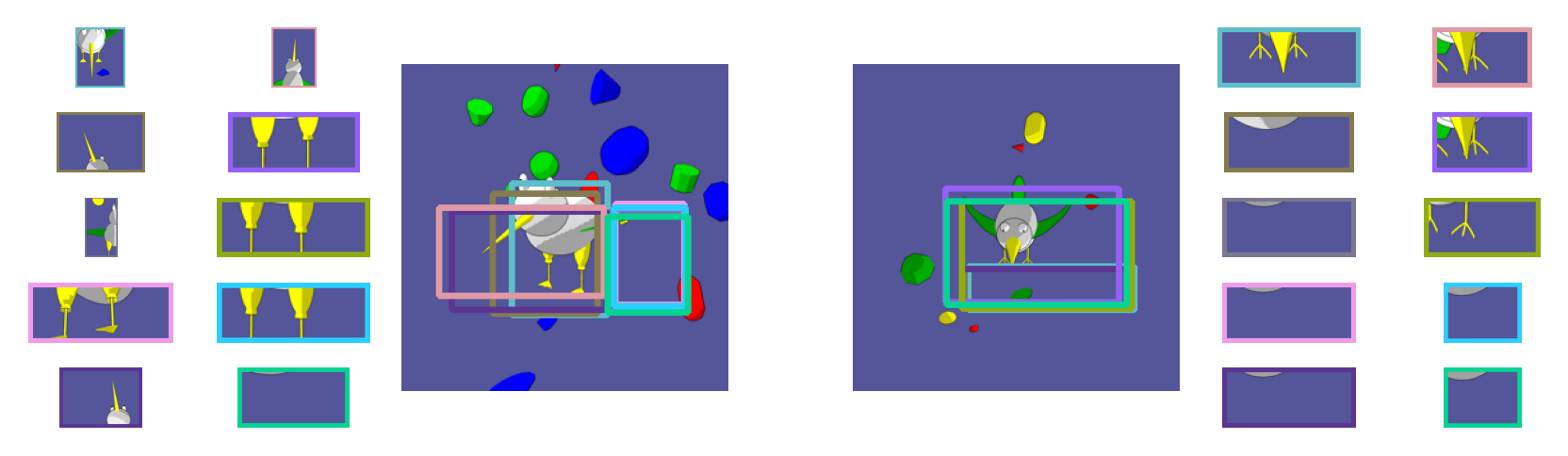}
   \vspace{-15pt}
   \caption{\emph{Qualitative results of ProtoPNet}~\cite{Chen:2019:TLL}. Many of the test image patches (center) highlight the entire bird or just background. Further, prototypes (sides) often do not include important parts.}
   \label{fig:results_pp}
   \vspace{-0.8em}
\end{figure}

\myparagraphnospace{(7)} ProtoPNet~\cite{Chen:2019:TLL} has poor correctness and contrastivity,
which can be explained by related findings in that ProtoPNet similarities are not semantically reasonable~\cite{Kim:2022:EHI} and by its poor localization (see \cref{sec:custom_eval} and \cref{fig:results_pp}).

\myparagraphnospace{(8)} Especially \emph{correctness} is a weakness of all examined methods, showing that, while existing approaches tend to accurately identify important areas, they lack the ability to reliably communicate the relative importance of each input area, or that the simple explanation of an attribution map might be too strong of a simplification of the actual model.

Finally, since XAI evaluation is multi-faceted and downstream task dependent, we doubt that a single score can be enough to properly compare methods. Hence, our work is less about ranking but more about automatically revealing the strengths and weaknesses of existing and future XAI methods in a systematic and theoretically grounded manner.

\subsection{Custom evaluations}\label{sec:custom_eval}

\paragraph{ProtoPNet.} To quantitatively evaluate if prototypes used in \cite{Chen:2019:TLL} promote explainability (see \cref{sec:evaluation:custom}), we conduct several sanity checks. First, we measure that out of $1000$ examined ProtoPNet patches in the test image, only $510$ touch the same part in the corresponding prototype patch, and thus, have a similarity that could be aligned with human notions of similarity. Further, we measure that $26.3\%$ of the extracted test image patches do not touch a single bird part, indicating a poor localization of the explanation. Finally, $18.5\%$ of the test image patches entail $4$ or $5$ different bird parts, which shows that the extracted patches are not local but global, and thus, do not promote the interpretability of the examined model. Qualitative results showing the described issues can be seen in \cref{fig:results_pp}. 
To summarize, many of the communicated prototypes of ProtoPNet~\cite{Chen:2019:TLL} trained on our dataset are unlikely to be meaningful for humans. 
While a similar conclusion was drawn in a recent \emph{human} study~\cite{Kim:2022:EHI} confirming our findings, our proposed tools allow the same conclusion in a fully \emph{automatic} manner. 

\begin{figure}[t]
  \centering
   \includegraphics[width=1\linewidth]{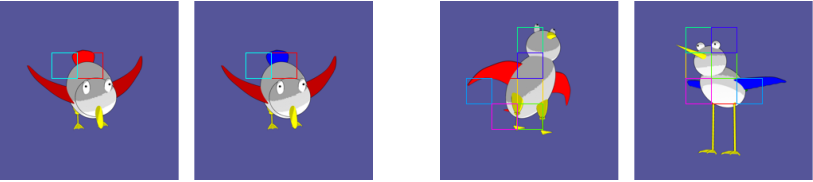}
   \vspace{-13pt}
   \caption{\emph{Qualitative results of counterfactual visual explanations (CVE)}~\cite{Goyal:2019:CVE}\emph{.} \emph{(left)} CVE for an aligned image pair with one part being different. \emph{(right)} CVE for an unaligned image pair with multiple parts being different. Note how many of the proposed swaps are unrealistic, and therefore, of little quality (\eg, the purple patch around the right eye is swapped with the neck of the other bird).}
   \label{fig:results_cve}
   \vspace{-0.8em}
\end{figure}

\myparagraph{Counterfactual visual explanations.} To quantify if CVE~\cite{Goyal:2019:CVE} highlights meaningful swaps (see \cref{sec:evaluation:custom}), we measure how many of the extracted swaps do not touch the same bird parts in the two images. When considering images that have the same viewpoint and differ only by one part (\cref{fig:results_cve} -- left), CVE with VGG16\ \cite{Simonyan:2015:VDC} produces on average $1.65$ swaps that do not touch the same part, and therefore, are semantically meaningless. When changing the viewpoint, CVE produces on average $1.3$ swaps that do not touch the same part. 
When working with aligned viewpoints and birds that differ by multiple parts ($4.07$ parts on average), $2.75$ such swaps are produced and $2.62$ when also changing the viewpoint (\cref{fig:results_cve} -- right). 
Interestingly, the number of different parts has a stronger impact on the absolute number of unreasonable swaps, \ie, swaps that do not touch the same part in both images, than the viewpoint. Further, we observe that even in our controlled environment, CVE produces on average $\sim 3$ unreasonable swaps for images with on average only $4$ different parts, indicating that many of the proposed swaps are semantically incorrect, and thus, do not promote coherent explanations.

\section{Conclusion}
In this work, we propose a novel approach to \emph{automatically} analyzing XAI methods using a synthetic classification dataset that allows for full annotation and part interventions. Using this dataset, we propose an accompanying \emph{multi-dimensional} analysis framework that \textit{faithfully} assesses various important aspects of explainability and generalizes to different explanation types by using \textit{interface functions}. With this easy-to-use tool, we analyzed 24 different setups to uncover various new insights and confirm findings from related work. This shows that, despite the synthetic setting, our findings appear to translate well to real data, and that our proposed tool is a practical and valuable asset for the analysis of future XAI methods. 
Finally, we showed how to develop tailored analyses to gain a better understanding of two specific XAI methods and discovered their weaknesses in an automatic and quantitative manner.

{\myparagraph{Acknowledgements.} We thank Bernt Schiele, Moritz Böhle, and Sukrut Rao for their valuable feedback and helpful discussions.
This project has received funding from the
European Research Council (ERC) under the European Union’s Horizon 2020 research and innovation programme (grant agreement No.~866008).
The project has also been
supported in part by the State of Hesse through the cluster projects “The Third Wave of Artificial
Intelligence (3AI)” and “The Adaptive Mind (TAM)”.}

{\small
\bibliographystyle{ieee_fullname}
\bibliography{bibtex/short,bibtex/papers,bibtex/external,bibtex/local}
}

\clearpage
\newpage

\appendix
\pagenumbering{roman}

\section{Overview}
\begin{figure*}[t]
  \centering
   \includegraphics[width=1.0\linewidth]{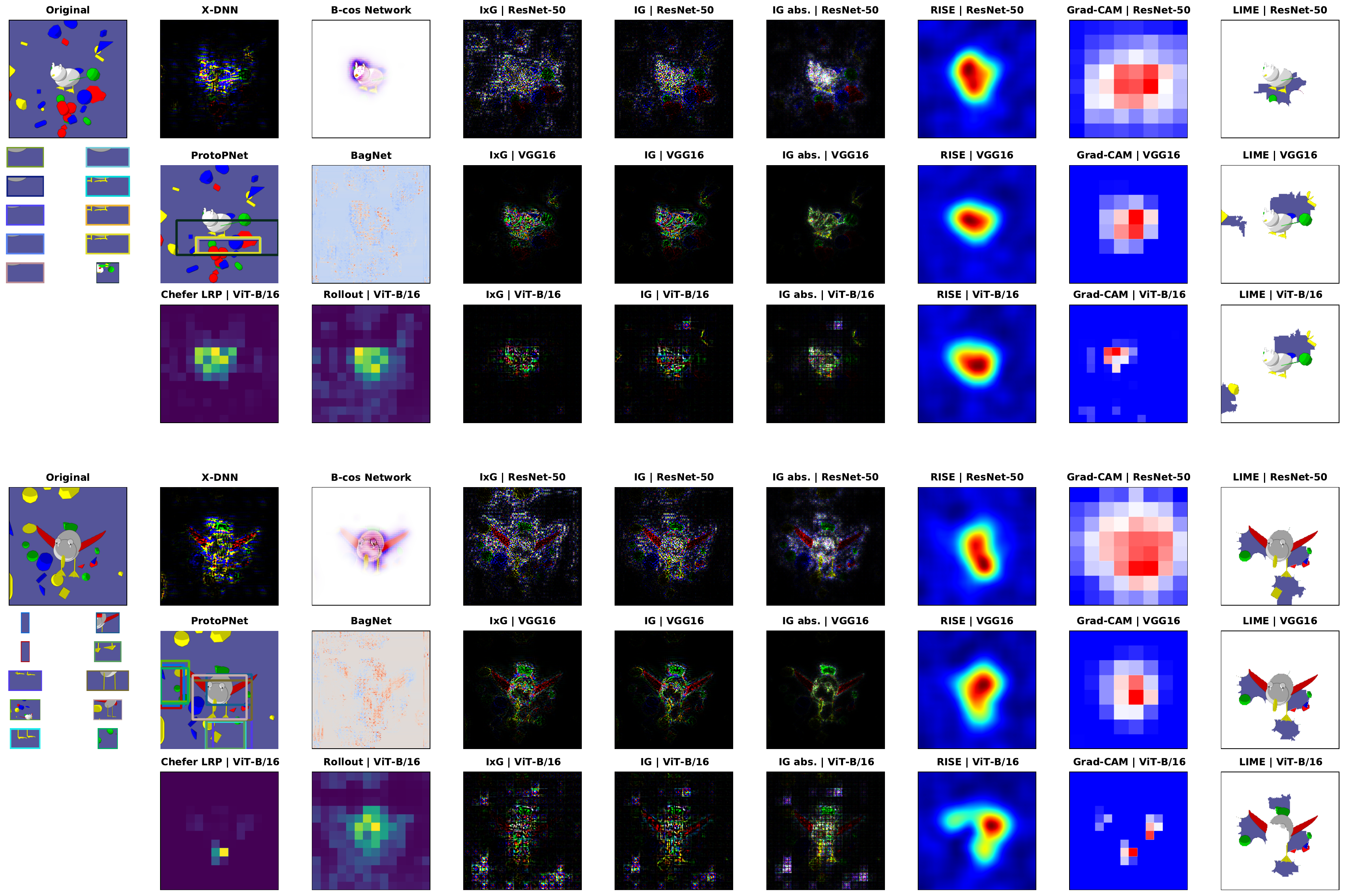}

   \caption{\emph{Additional qualitative results for the examined explanation methods.} Each group of three rows shows results for the same input image and all respective XAI methods and backbones that have been examined in our \emph{FunnyBirds} framework. The displayed qualitative results are consistent with the qualitative results in \cref{fig:results} from the main paper.}
   \label{fig:qualitative_res}
\end{figure*}

This appendix provides additional information, framework results (\cref{tab:protocols_sup}), qualitative results (\cref{fig:qualitative_res,fig:example_imgs}), and experimental details for reproducibility purposes, which could not be included in the main text due to space limitations.

\section{Interface Functions}

Our proposed interface functions have to be instantiated individually for each explanation type. In our evaluation, we examine four explanation types:
\begin{enumerate}
    \item  \emph{Attribution maps}, \ie, Input$\times$Gradient~\cite{Shrikumar:2016:NJB}, (absolute) Integrated Gradients~\cite{Sundararajan:2017:AAD}, Grad-CAM~\cite{Selvaraju_2017:GCV}, RISE~\cite{Petsiuk:2018:RIS}, $\mathcal{X}$-DNN~\cite{Hesse:2021:FAA}, BagNet~\cite{Brendel:2019:ACB}, B-cos networks~\cite{Boehle:2022:BCN}, and Chefer LRP~\cite{Chefer:2021:TIB}. 
    \item  \emph{Attention from vision transformers}~\cite{Dosovitskiy:2021:IWW}, \ie, Rollout~\cite{Abnar:2020:QAF}.
    \item \emph{Binary importance maps} that indicate if a pixel is important or not, \ie, LIME~\cite{Ribeiro:2016:WSI}. 
    \item \emph{Prototypes} that allow for explanations of the type “this looks like that,” \ie, ProtoPNet~\cite{Chen:2019:TLL}. 
\end{enumerate}
In the following, we will outline how interface functions for these explanation types are instantiated:

\paragraph{Attribution maps:} 
\begin{description}
  \item[$\textnormal{PI}(\cdot)\!$] -- The part importance is estimated by summing the pixelwise attribution scores within each part, where the part mask is dilated with a small square kernel of size $5 \times 5$ to also include the part's edges.
  \item[$\textnormal{P}(\cdot)\!$] --  A part is considered to be important if its part importance is more than $t\%$ of the total attribution sum of the explanation. For a description of how $t$ is chosen, please refer to \cref{sec:evaluation:com} of the main paper.
\end{description}

\begin{figure}[t]
  \centering
   \includegraphics[width=\linewidth]{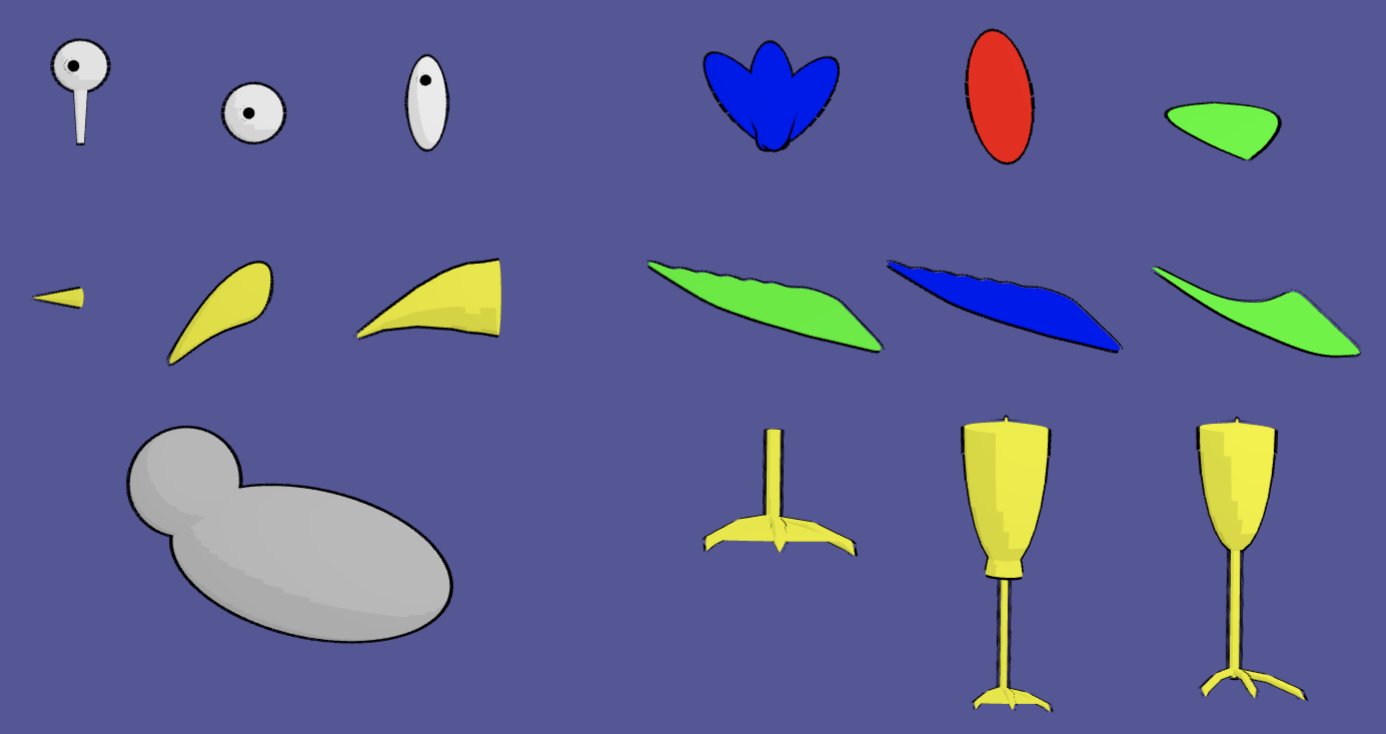}

   \caption{Example models of the neutral \textit{body} (bottom left) and our dataset concepts (\ie, parts) \textit{beak}, \textit{wings}, \textit{feet}, \textit{eyes}, and \textit{tail}.}
   \label{fig:parts}
\end{figure}

\paragraph{Attention:} 
\begin{description}
  \item[$\textnormal{PI}(\cdot)\!$] -- The part importance is estimated by summing the pixelwise attribution scores within each part, where the part mask is dilated with a small square kernel of size $5 \times 5$ to also include the part's edges. Since attention rollout~\cite{Abnar:2020:QAF} cannot be computed \wrt a particular target class~\cite{Chefer:2021:TIB}, the target sensitivity protocol cannot be computed.
  \item[$\textnormal{P}(\cdot)\!$] --  A part is considered to be important if its part importance is more than $t\%$ of the total attribution sum of the explanation.
\end{description}

\paragraph{Binary importance maps:} 
\begin{description}
  \item[$\textnormal{PI}(\cdot)\!$] -- The part importance is not defined for this explanation type.
  \item[$\textnormal{P}(\cdot)\!$] --  A part is considered to be important if $t\%$ of its pixels are estimated to be important by the explanation.
\end{description}

\begin{figure*}[t]
  \centering
   \includegraphics[width=1.0\linewidth]{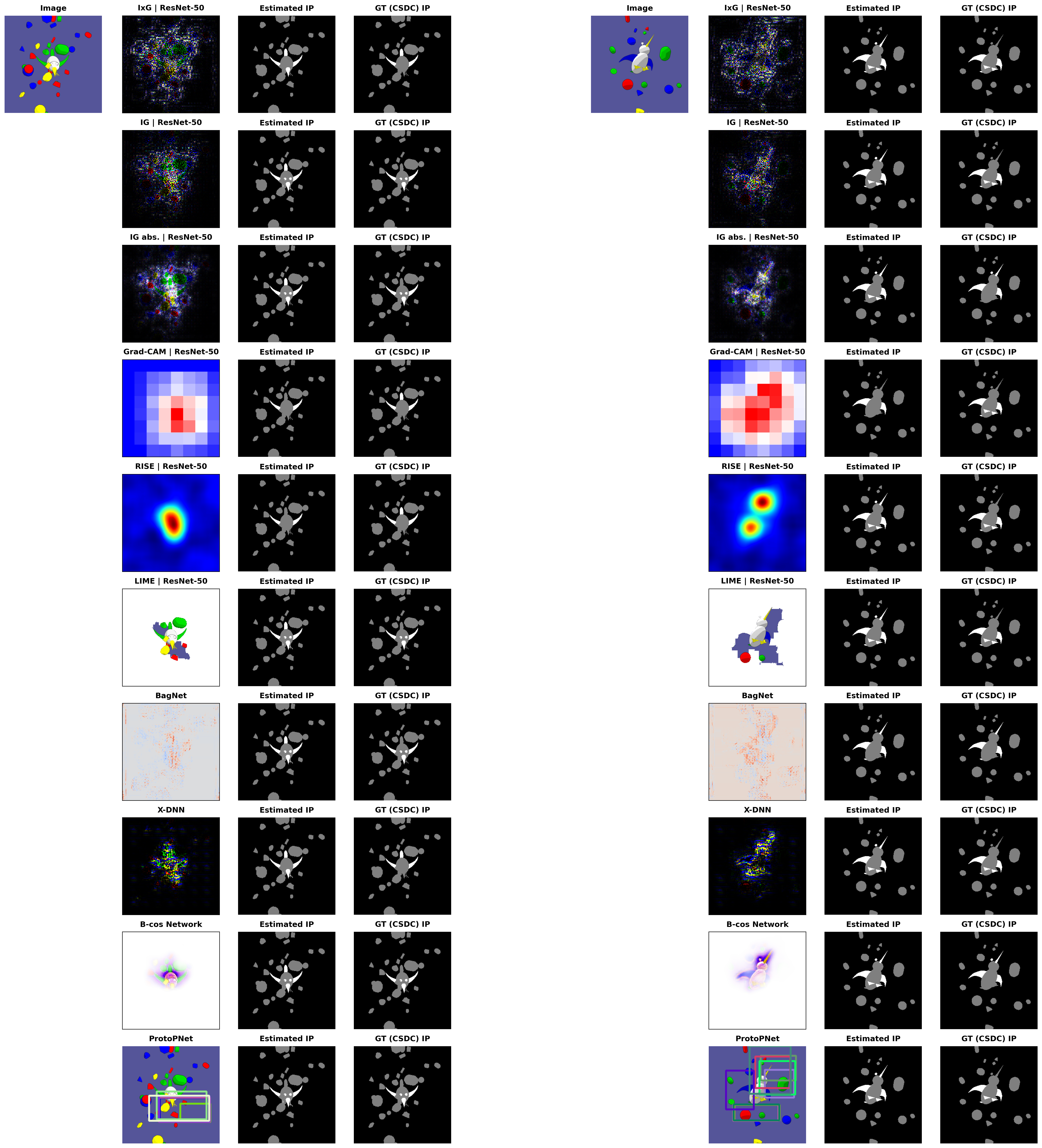}
   \caption{\emph{Explanations and the extracted important parts.} The left column of each block shows the original input image. Next, we show the explanation from each respective XAI method. This is followed by the estimated important parts (estimated IP -- highlighted in white) from the explanation using our interface function $\text{P}(\cdot)$ with a threshold $t=0.02$. In the last column, we show the ground-truth minimal important parts from the \textit{controlled synthetic data check} protocol (GT (CSDC) IP). For example, the parts estimated to be important by Grad-CAM~\cite{Selvaraju_2017:GCV} in the left block are not fully complete, since fewer parts are highlighted than for GT (CSDC) IP.
   }
   \label{fig:important_parts_viz}
\end{figure*}

\paragraph{Prototypes:} 
\begin{description}
  \item[$\textnormal{PI}(\cdot)\!$] -- The importance score within the bounding box of a single prototype is the product of its similarity score and its class connection score. The importance score outside of the bounding box is zero. The final part importance is estimated by summing the importance scores of each prototype belonging to the class of interest.
  \item[$\textnormal{P}(\cdot)\!$] --  A part is considered to be important if $t\%$ of its pixels overlap with the explanation's bounding boxes.
\end{description}
Qualitative results for $\text{P}(\cdot)$ can be seen in \cref{fig:important_parts_viz}.

\begin{table*}
  \centering
  \caption{\textit{Quantitative results of our FunnyBirds evaluation protocols.} See \cref{sec:evaluation:benchmark} of the main paper for a description of the evaluation metrics. We additionally report the final scores for each respective explainability dimension completeness (Com.), correctness (Cor.), and contrastivity (Con.). The mean explainability score (mX) denotes the mean of the final completeness, correctness, and contrastivity scores. Note that A (Acc.), BI (B.I.), Com., Cor., Con., and mX are also included in \cref{fig:results} of the main paper. * denotes slight architectural changes.}
  \smallskip
  \small
  \begin{tabularx}{1.0\textwidth}{@{}llllllllllllll@{}}
    \toprule
    \textbf{Backbone} & \textbf{XAI Method}      & \textbf{A} & \textbf{BI} & \textbf{CSDC} & \textbf{PC} & \textbf{DC} & \textbf{D} & \textbf{SD} & \textbf{TS}& \textbf{Com.} & \textbf{Cor.} & \textbf{Con.} & \textbf{mX}\\
    \midrule
VGG16~\cite{Simonyan:2015:VDC} & IG~\cite{Sundararajan:2017:AAD} & 0.99&0.99&0.92&0.92&0.92&0.97&0.67&0.92&0.95&0.67&0.92&0.85\\
VGG16~\cite{Simonyan:2015:VDC} & IG abs.~\cite{Sundararajan:2017:AAD} & 0.99&0.99&0.96&0.99&0.97&0.97&0.69&0.84&0.97&0.69&0.84&0.83\\
VGG16~\cite{Simonyan:2015:VDC} & RISE~\cite{Petsiuk:2018:RIS} & 0.99&0.99&0.8&0.73&0.7&0.84&0.73&0.83&0.79&0.73&0.83&0.78\\
VGG16~\cite{Simonyan:2015:VDC} & LIME~\cite{Ribeiro:2016:WSI} & 0.99&0.99&0.89&0.88&0.9&0.92&0&0&0.91&0&0&0.3\\
VGG16~\cite{Simonyan:2015:VDC} & IxG~\cite{Shrikumar:2016:NJB} & 0.99&0.99&0.79&0.71&0.69&0.94&0.55&0.69&0.84&0.55&0.69&0.69\\
VGG16~\cite{Simonyan:2015:VDC} & Grad-CAM~\cite{Selvaraju_2017:GCV} & 0.99&0.99&0.94&0.97&0.93&0.87&0.75&0.93&0.91&0.75&0.93&0.86\\\midrule
ResNet-50~\cite{He:2016:DRL} & IG~\cite{Sundararajan:2017:AAD} & 1&1&0.92&0.94&0.88&0.81&0.59&0.98&0.86&0.59&0.98&0.81\\
ResNet-50~\cite{He:2016:DRL} & IG abs.~\cite{Sundararajan:2017:AAD} & 1&1&0.95&0.97&0.91&0.79&0.53&0.86&0.87&0.53&0.86&0.75\\
ResNet-50~\cite{He:2016:DRL} & RISE~\cite{Petsiuk:2018:RIS} & 1&1&0.82&0.75&0.74&0.63&0.56&0.61&0.7&0.56&0.61&0.62\\
ResNet-50~\cite{He:2016:DRL} & LIME~\cite{Ribeiro:2016:WSI} & 1&1&0.94&0.94&0.92&0.78&0&0&0.86&0&0&0.29\\
ResNet-50~\cite{He:2016:DRL} & IxG~\cite{Shrikumar:2016:NJB} & 1&1&0.74&0.61&0.53&0.54&0.54&0.8&0.58&0.54&0.8&0.64\\
ResNet-50~\cite{He:2016:DRL} & Grad-CAM~\cite{Selvaraju_2017:GCV} & 1&1&0.8&0.74&0.69&0.74&0.55&0.78&0.74&0.55&0.78&0.69\\
\midrule
ViT-B/16~\cite{Dosovitskiy:2021:IWW} & IG~\cite{Sundararajan:2017:AAD} & 0.98&1&0.89&0.86&0.85&0.9&0.65&0.91&0.88&0.65&0.91&0.82\\
ViT-B/16~\cite{Dosovitskiy:2021:IWW} & IG abs.~\cite{Sundararajan:2017:AAD} & 0.98&1&0.96&0.98&0.95&0.89&0.63&0.74&0.92&0.63&0.74&0.76\\
ViT-B/16~\cite{Dosovitskiy:2021:IWW} & RISE~\cite{Petsiuk:2018:RIS} & 0.98&1&0.79&0.71&0.7&0.83&0.79&0.75&0.78&0.79&0.75&0.77\\
ViT-B/16~\cite{Dosovitskiy:2021:IWW} & LIME~\cite{Ribeiro:2016:WSI} & 0.98&1&0.95&0.96&0.96&0.85&0&0&0.9&0&0&0.3\\
ViT-B/16~\cite{Dosovitskiy:2021:IWW} & IxG~\cite{Shrikumar:2016:NJB} & 0.98&1&0.74&0.59&0.6&0.43&0.51&0.67&0.54&0.51&0.67&0.57\\
ViT-B/16~\cite{Dosovitskiy:2021:IWW} & Grad-CAM~\cite{Selvaraju_2017:GCV, Chefer:2021:TIB} & 0.98&1&0.75&0.67&0.68&0.91&0.7&0.48&0.81&0.7&0.48&0.66\\
ViT-B/16~\cite{Dosovitskiy:2021:IWW} & Rollout~\cite{Abnar:2020:QAF} & 0.98&1&0.86&0.8&0.82&0.8&0.76&0&0.81&0.76&0&0.52\\
ViT-B/16~\cite{Dosovitskiy:2021:IWW} & Chefer LRP~\cite{Chefer:2021:TIB} & 0.98&1&0.91&0.92&0.89&0.9&0.74&0.95&0.9&0.74&0.95&0.86\\
\midrule
BagNet~\cite{Brendel:2019:ACB} & BagNet~\cite{Brendel:2019:ACB} & 1&1&0.95&0.98&0.91&0.91&0.76&0.99&0.93&0.76&0.99&0.9\\
ResNet-50*~\cite{He:2016:DRL} & B-cos network~\cite{Boehle:2022:BCN} & 0.96&0.87&0.93&0.88&0.94&0.86&0.69&0.89&0.89&0.69&0.89&0.82\\
ResNet-50*~\cite{He:2016:DRL} & $\mathcal{X}$-DNN~\cite{Hesse:2021:FAA} & 0.99&1&0.9&0.88&0.85&0.93&0.6&0.87&0.91&0.6&0.87&0.79\\
ResNet-50~\cite{He:2016:DRL} & ProtoPNet~\cite{Chen:2019:TLL} & 0.94&1&0.93&0.91&0.92&0.58&0.24&0.46&0.75&0.24&0.46&0.48\\
    \bottomrule
  \end{tabularx}
  \label{tab:protocols_sup}
\end{table*}

\section{Evaluation Details}
\subsection{Accuracy and background independence}

In the main paper, we describe the accuracy ($\text{A}$) and background independence ($\text{BI}$) metrics only textually. We here provide the accompanying formulas. The notation follows that of \cref{sec:evaluation:benchmark} in the main paper.

The accuracy $\text{A}$ denotes the standard classification accuracy:
\begin{equation}
\text{A} = \frac{1}{N} \sum_{n=1}^{N} \big[f(x_n) = c_n \big]\ ,
\end{equation}
with $[\cdot]$ denoting the Iverson bracket~\cite{Knuth:1992:TNN}. 

The background independence $\text{BI}$ denotes the ratio of background objects that, when removed, cause the model output to drop less than $5\%$:
\begin{equation}
\text{BI} = \frac{1}{NB} \sum_{n=1}^{N} \big[|0.05 f(x_n)| > f(x_{n}) - \{f(x_{n}^{\prime\prime\prime\prime})\} \big]\ ,
\end{equation}
with $B$ denoting the average number of background objects per image, and $\{f(x_{n}^{\prime\prime\prime\prime})\}$ the target scores of the images obtained by removing any individual background object from image $x_n$. 

\subsection{Other dimensions of evaluation}

In our FunnyBirds framework, we analyze the explainability dimensions \textit{completeness}, \textit{correctness}, and \textit{contrastivity}. With our custom evaluations we evaluate the \textit{coherence} of methods. 
Nauta \etal~\cite{Nauta:2022:FAE} propose various additional dimensions for evaluating XAI that have been studied in related work but are not considered in our paper. 
We here describe the reasons for not including these dimensions; please refer to \cite{Nauta:2022:FAE} for a definition of each dimension:

\begin{description}
\item[Consistency.] Consistency received only minor attention in related work.
\item[Continuity.] Continuity received only moderate attention in related work. Additionally, we believe that the continuity of an explanation is usually strongly tied to the continuity of the model. Nevertheless, continuity could be included in our framework by measuring the part importance differences for two similar input images, \eg, two images with slight viewpoint or illumination changes.
We leave this for future work.

\item[Compactness.] We believe that there is a discrepancy between the automatically measurable size of an explanation and the size of an explanation that is perceived by a human. For example, an attribution map contains a lot of information (in the sense of memory size, \eg, in MiB) that, however, is much easier for a human to parse than, \eg, a complex mathematical function that requires only little memory to store. For this reason, we do not believe that compactness of different explanation types can be sufficiently evaluated without a human in the loop.

\item[Covariate complexity.] Covariate complexity received only moderate attention in related work. Additionally, just as compactness, it is strongly related to a human assessment, and thus, not qualified for our fully automatic framework. However, one could develop custom evaluations that measure the covariate complexity of specific methods.

\item[Composition.] Composition received only minor attention in related work.

\item[Confidence.] Confidence received only minor attention in related work.

\item[Context.] Context received only minor attention in related work.

\item[Controllability.] Controllability received only minor attention in related work.
\end{description}

\section{Experimental Details}

All examined models are initialized with weights obtained by pre-training on ImageNet~\citelatex{Deng:2009:ILS}. If not specified otherwise, we use the hyper-parameters from the original implementation of each respective model. To ensure that test images with removed parts are from the same distribution as the training data, we augment half of our training set by randomly removing $n \in \{0,\ldots,5\}$ bird parts from each image. Since these images can no longer be distinctly associated with one specific class, we utilize a multi-label classification training scheme, where we compute the average cross-entropy loss for all potential targets, \ie, all the classes that contain all the remaining parts. For training, we use a single NVIDIA A100-SXM4-40GB GPU. We train each model twice and select the run with the higher test set accuracy for our evaluation.

\myparagraph{ResNet-50.} To train ResNet-50~\cite{He:2016:DRL}, we use a batch size of 64 and an SGD optimizer with a weight-decay of $1e{-}4$, a momentum of $0.9$, and a learning rate of $0.1$; we train for $120$ epochs with a learning rate scheduler that multiplies the initial learning rate with a factor of $0.1$ after $60$ epochs.

\myparagraph{VGG16.} To train the VGG16~\cite{Simonyan:2015:VDC} model, we use the same training setup as for the ResNet-50 model with an initial learning rate of $0.001$.

\myparagraph{ViT-B/16.} To train the ViT-B/16~\cite{Dosovitskiy:2021:IWW} model, we use the same training setup as for the ResNet-50 model with an initial learning rate of $0.01$.

\myparagraph{$\mathcal{X}$-DNN.} To train the $\mathcal{X}$-DNN~\cite{Hesse:2021:FAA} model, we use the same training setup as for the ResNet-50 model with an initial learning rate of $0.01$.

\myparagraph{BagNet.} To train the BagNet~\cite{Brendel:2019:ACB} model, we use the same training setup as for the ResNet-50 model with an initial learning rate of $0.01$. Our instantiation of BagNet uses a receptive field of $33 \times 33$.

\myparagraph{B-cos network.} To train the B-cos network~\cite{Boehle:2022:BCN}, we use the same training setup as for the ResNet-50 model with an initial learning rate of $0.01$. As recommended in the original paper~\cite{Boehle:2022:BCN}, we use a binary cross-entropy loss, and we concatenate the input $x'$ with its complement, giving us the final input $x = [x',1-x']$.

\myparagraph{ProtoPNet.} To train ProtoPNet~\cite{Chen:2019:TLL}, we use the same hyper-parameters as in the original paper, \ie, a batch size of 80, a learning rate of $1e{-}4$ for the features and $3e{-}3$ for the add-on layers and prototype vectors, $100$ training epochs, and a learning rate decay factor of $0.1$ after every 5 epochs.

\subsection{Dataset generation} 
\begin{figure*}[t]
  \centering
   \includegraphics[width=1.0\linewidth]{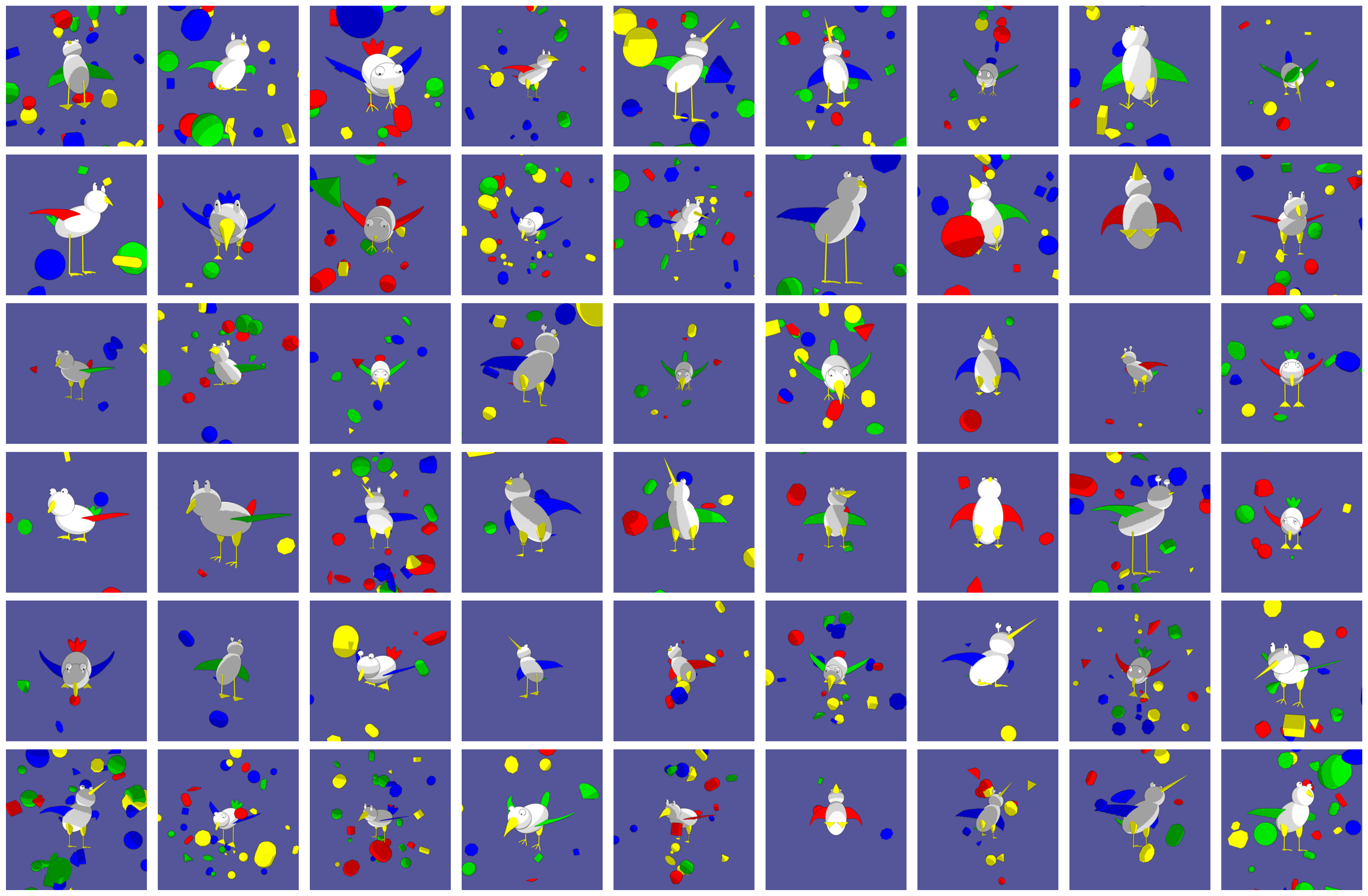}

   \caption{\emph{Example images from our FunnyBirds dataset.}}
   \label{fig:example_imgs}
\end{figure*}

Our proposed dataset consists of rendered $3$D scenes, as shown in \cref{fig:example_imgs}. The required bird parts are manually modeled using Blender.\footnote{\href{http://www.blender.org/}{blender.org}} To render the scenes we use Three.js, a JavaScript $3$D Library.\footnote{\href{http://www.threejs.org/}{threejs.org}} For our proposed toon shading, we use \emph{MeshToonMaterial}.\footnote{\href{https://threejs.org/docs/\#api/en/materials/MeshToonMaterial}{threejs.org -- MeshToonMaterial}} In order to add shadows and achieve a 3D effect, we add a point light source to the scene. 
We empirically validated that an image with all bird parts removed cannot be classified beyond random guessing, to ensure that the background contains \textit{no} class-specific information.

\subsection{Stability across runs} \label{sec:stability}

\begin{table*}
  \centering
  \caption{\textit{Stability of our evaluation protocols.} Scores indicate the absolute difference of two runs. See \cref{sec:evaluation:benchmark} of the main paper for a description of the metrics. The mean explainability score (mX) denotes the mean of the final completeness, correctness, and contrastivity scores. We report results for the stability across \textit{(1)} two training runs and \textit{(2)} between the original and a larger test set (see \cref{sec:stability}).}
  \smallskip
  \small
  \begin{tabularx}{1.0\textwidth}{llXlllllllll}
    \toprule
    \textbf{Setup} & \textbf{Backbone} & \textbf{XAI Method} & \textbf{$|\Delta \textbf{A}|$} & $|\Delta \textbf{BI}|$ & $|\Delta \textbf{CSDC}|$ & $|\Delta \textbf{PC}|$ & $|\Delta \textbf{DC}|$ & $|\Delta \textbf{D}|$ & $|\Delta \textbf{SD}|$ & $|\Delta \textbf{TS}|$ & $|\Delta \textbf{mX}|$\\
    \midrule
(1) & VGG16~\cite{Simonyan:2015:VDC} & IG~\cite{Sundararajan:2017:AAD}& 0.002&0.003&0.008&0.008&0.008&0.001&0.015&0.005&0.005\\
(1) & ResNet-50~\cite{He:2016:DRL} & IG~\cite{Sundararajan:2017:AAD} & 0&0.001&0&0.02&0.006&0&0.022&0.002&0.008\\
(1) & ViT-B/16~\cite{Dosovitskiy:2021:IWW} & IG~\cite{Sundararajan:2017:AAD} & 0.01&0&0.051&0.11&0.104&0.016&0.031&0.037&0.035\\
(1) & VGG16~\cite{Simonyan:2015:VDC} & IxG~\cite{Simonyan:2015:VDC}& 0.002&0.003&0.005&0.002&0.066&0.012&0.019&0.071&0.023\\
(1) & ResNet-50~\cite{He:2016:DRL} & IxG~\cite{Simonyan:2015:VDC} & 0&0.001&0.02&0.012&0.032&0.031&0.003&0.015&0.003\\
(1) & ViT-B/16~\cite{Dosovitskiy:2021:IWW} & IxG~\cite{Simonyan:2015:VDC} & 0.01&0&0.066&0.12&0.126&0.026&0.017&0.06&0.039\\
\midrule
(2) & VGG16~\cite{Simonyan:2015:VDC} & IG~\cite{Sundararajan:2017:AAD} & 0.0128&0.001&0.008&0.027&0.004&0.013&0&0.007&0.001\\
(2) & ResNet-50~\cite{He:2016:DRL} & IG~\cite{Sundararajan:2017:AAD} & 0.0124&0&0.008&0.079&0.003&0.029&0.032&0.005&0.009\\
(2) & ViT-B/16~\cite{Dosovitskiy:2021:IWW} & IG~\cite{Sundararajan:2017:AAD} & 0.0232&0&0.008&0.082&0.02&0.017&0.016&0.008&0.006\\
(2) & VGG16~\cite{Simonyan:2015:VDC} & IxG~\cite{Simonyan:2015:VDC} & 0.0128&0.001&0.004&0.054&0.001&0&0.006&0.013&0.001\\
(2) & ResNet-50~\cite{He:2016:DRL} & IxG~\cite{Simonyan:2015:VDC} & 0.0124&0&0.011&0.079&0.03&0.005&0.009&0.005&0.007\\
(2) & ViT-B/16~\cite{Dosovitskiy:2021:IWW} & IxG~\cite{Simonyan:2015:VDC} & 0.0232&0&0.039&0.106&0.053&0.058&0.004&0&0.003 \\
    \bottomrule
  \end{tabularx}
  \label{tab:std}
\end{table*}

To measure the stability of our FunnyBirds framework across runs, we report the absolute difference of two runs in \cref{tab:std}.
We report results for two setups: \textit{(1)} the absolute difference between the evaluation on two training runs (\ie, trained with differing random seeds) and \textit{(2)} the absolute difference between evaluating the respective model from the main paper on the original test set (500 samples) and a larger test set with 2\,500 samples. 
This allows us to measure the stability across different training runs and across different test set sizes. 
The absolute difference between different training runs is fairly small ($\leq 0.039$ for mX, see \cref{tab:std}). 
CNN-based architectures appear to be more stable than the vision transformer. 
Also, the absolute difference of the explainability protocols across runs is somewhat correlated with the absolute difference of the accuracy across runs. 
This suggests that the fluctuation of the accuracy across different training runs is a good proxy for the stability of the explanation protocols. 
This may be due to models with similar accuracy learning similar functions, and thus, providing similar explanations. 

The absolute difference between evaluating on the original test set and on a larger test set is even smaller ($\leq 0.009$ for mX), indicating that the proposed dataset size (500 images) is sufficiently large.
We purposely did not use the larger test set for the principal evaluation in the main paper to keep the computational expense at bay and allow for an easy adoption of our analysis framework in future work. 
Note that for slower explanation methods like RISE~\cite{Petsiuk:2018:RIS}, evaluating 2\,500 images would take $\sim50$h on an NVIDIA A100-SXM4-40GB GPU, which would impair the practicability of our proposed framework. 
Nevertheless, we will also publish the larger test set for evaluation under these conditions.
To conclude, we find that our framework is quite stable under different training runs and that our test set size is sufficiently large.


{\small

\bibliographystylelatex{ieee_fullname}
\bibliographylatex{bibtex/short,bibtex/papers,bibtex/external,bibtex/local}
}

\end{document}